\documentclass[sigconf]{acmart}

\usepackage{subfigure}
\usepackage{multirow}
\usepackage{multicol}
\usepackage{fancyhdr}

\AtBeginDocument{%
  \providecommand\BibTeX{{%
    \normalfont B\kern-0.5em{\scshape i\kern-0.25em b}\kern-0.8em\TeX}}}

\setcopyright{acmcopyright}
\copyrightyear{2022}
\acmYear{2022}
\acmDOI{XXXXXXX.XXXXXXX}

\acmConference[MM '22]{Proceedings of the 30th ACM International Conference on Multimedia}{October 10--14, 2022}{Portugal, Lisbon}
\acmBooktitle{Proceedings of the 30th ACM International Conference on Multimedia (MM '22), October 10--14, 2022, Portugal, Lisbon}
\acmPrice{15.00}
\acmISBN{978-1-4503-XXXX-X/18/06}

\acmConference[Lisbon '22]{Lisbon '22: ACM Multimedia}{Oct 10--14, 2022}{Lisbon, Portugal}

\acmSubmissionID{21}

\begin{document}

\title{Benign Adversarial Attack: Tricking Models for Goodness}	

%

\author{Jitao Sang, Xian Zhao, Jiaming Zhang, Zhiyu Lin}
\affiliation{%
  \institution{School of Computer and Information Technology \& Beijing Key Lab of Traffic Data Analysis and Mining,\\
 Beijing Jiaotong University}
  \city{Beijing}
  \country{China}
}
\email{{jtsang, xianzhao, jiamingzhang, zhiyulin}@bjtu.edu.cn}


\begin{abstract}
In spite of the successful application in many fields, machine learning models today suffer from notorious problems like vulnerability to adversarial examples. Beyond falling into the cat-and-mouse game between adversarial attack and defense, this paper provides alternative perspective to consider adversarial example and explore whether we can exploit it in benign applications. We first attribute adversarial example to the human-model disparity on employing non-semantic features. While largely ignored in classical machine learning mechanisms, non-semantic feature enjoys three interesting characteristics as (1) exclusive to model, (2) critical to affect inference, and (3) utilizable as features. Inspired by this, we present brave new idea of benign adversarial attack to exploit adversarial examples for goodness in three directions: (1) adversarial Turing test, (2) rejecting malicious model application, and (3) adversarial data augmentation. Each direction is positioned with motivation elaboration, justification analysis and prototype applications to showcase its potential.

\end{abstract}




\maketitle

\pagestyle{empty}

\section{Introduction}
Ever since its discovery, adversarial examples have been exploited to attack machine learning models ranging from perception tasks like landmark recognition~\cite{duan2021adversarial}, pedestrian detection~\cite{xu2020adversarial}, face recognition~\cite{sharif2016accessorize}, to cognition tasks like decision making~\cite{su2019one}, and even system breakdown~\cite{huang2017adversarial}.  Due to the critical risk to machine learning algorithms, adversarial attack has been viewed as malignant in default, which naturally gives rise to the considerable attention on circumventing adversarial examples. Recent years have therefore witnessed the iterative and endless ``involution'' between adversarial defense and attack solutions: from FGSM~\cite{goodfellow2014explaining}, C\&W attack~\cite{carlini2017towards}, PGD~\cite{madry2017towards} to Adversarial Transformation Networks~\cite{baluja2017adversarial} regarding adversarial attack, and from adversarial training~\cite{goodfellow2014explaining}, defensive distillation~\cite{papernot2016distillation}, gradient masking~\cite{papernot2017practical}, to DeepCloak~\cite{gao2017deepcloak} regarding adversarial defense. Once there is a stronger adversarial attack solution, a more robust adversarial defense solution is expected to be proposed accordingly (and vice versa). Other than falling into this cat-and-mouse game, this paper provides alternative perspective to consider adversarial example and explore whether we can exploit it in other-than-malignant applications.

We start with the discussion on why adversarial example exists. The hypothesis is that adversarial example results from the fact that model~\footnote{For the sake of simplicity, when no ambiguity is caused, we use ``model'' to denote ``machine learning model''.} is capable of employing information that human cannot well capture and understand, which we denote as \emph{non-semantic feature}. Using image as example, while human recognizes image relying mainly on the semantic information like shape and contour, machine learning models can harness additional non-semantic features imperceptible to humans to assist inference. Adversarial example, in this respect, corrupts the non-semantic features (with added adversarial perturbation) to mislead model inference result without affecting human perception. From this perspective, we look on adversarial example as one significant instantiation of non-semantic features, which demonstrate three interesting characteristics as: (1) exclusive to model, (2) critical to affect inference, and (3) utilizable as features.

Based on this alternative understanding and the associated characteristics, this paper then presents new idea to exploit adversarial attack for goodness, which we call \emph{benign adversarial attack}. Specifically, a framework is introduced with the following research directions corresponding to the three characteristics: (1) \emph{Adversarial Turing test}, employing the different sensitivities to adversarial perturbation to distinguish between human and model;  (2) \emph{Rejecting malicious model application}, specially designing adversarial attack to invalid malicious model applications like face recognition-based privacy violation; (3) \emph{Adversarial data augmentation}, discovering and harnessing more non-semantic features from the generated pseudo samples to address the data shortage problem. For each direction, we report preliminary experimental results for justification as well as providing prototype applications for illustration.

In the rest of the paper, Section~\ref{sec:2} attributes the existence of adversarial example to model's exclusive capability in employing non-semantic features and introduces three important characteristics of the non-semantic features. In Section~\ref{sec:3}, we present a framework of benign adversarial attack to exploit the adversarial examples for goodness. Three research directions are positioned with motivation, justification experiments and prototype applications. Finally, Section~\ref{sec:4} concludes the paper with discussions on balancing the trade-off between the risks and opportunities in employing non-semantic features.

\section{Adversarial Example: Exploiting the Human-Model Disparity}\label{sec:2}
\subsection{Human \emph{v.s.} Model: the Different Capabilities in Employing Non-semantic Feature}
In typical machine learning process, model learns knowledge from the training data annotated by human. Therefore, the training of machine learning model can be considered as distilling knowledge from human. In this case, the information employed by the model is expected to be consistent with human. However, recent studies suggest that human and model employ quite different information for inference. For example, ~\citet{wang2020high} first demonstrated that CNN can accurately recognize images only based on high-frequency components which are almost imperceptible to human. ~\cite{jere2020singular} claimed that CNN is sensitive to human-imperceptible higher-rank features from SVD and theoretically proved the relationship between high-frequency and high-rank features. Regarding adversarial example, a seminal study found that the ``non-robust features'' introduced from adversarial attack contribute remarkably to model's classification accuracy~\cite{ilyas2019adversarial}. 

Note that there is as yet no consensus on whether these human-imperceptible information are aggregations of random noise or real, predictive features of the task to be solved~\cite{buckner2020understanding}. We conduct this study and introduce the new idea on the basis of defending that these information provide useful and generalizable features. One important evidence is the transferability of adversarial examples among different model architectures and datasets~\cite{liu2016delving}. While human can hardly perceive and utilize these information, these information indeed encode the inherent data patterns and can be well utilized by models. 

An analog can be made with the primate trichromacy vision to help understand this difference between human and model. For most primate animals, they have three independent color receptors and a typical visible spectrum of 400–700 nm~\cite{nathans1994eye}. While, fish, birds and reptiles are tetrachromats which are also sensitive to light below 400 nm and can make use of ultraviolet (UV) information~\cite{marshall2014unconventional}. In this respect, we may treat machine learning model as a new species which is capable of processing human-imperceptible information.

We call these human-imperceptible information as \emph{non-semantic feature} that well generalizes to unseen data and is intrinsic to solve the target task. Reviewing the previous studies on discussing the human-model disparity, we divide non-semantic features into two categories: (1) weak-structured information from the perspective of content, such as high-frequency and higher-rank features~\cite{wang2020high,jere2020singular}, and (2) non-robust features from the perspective of model knowledge, which corresponds to the adversarial perturbation generated by attacking models~\cite{ilyas2019adversarial}.  

\subsection{Three Characteristics of Non-semantic Feature}
The above discussion suggests that non-semantic feature indeed contributes to solving the target task. However, its exploitation for applications is rarely seriously discussed. We now summarize three characteristics of non-semantic feature to motivate exploitation:
\begin{itemize}
\item \emph{Exclusive to model}: Solving tasks involves with both human-perceptible and human-imperceptible information. While machine learning models can utilize both, human relies mostly on the former part. Capability of processing non-semantic feature is the fundamental characteristic to distinguish between human and model.
\item \emph{Critical to affect inference}: Equipping with this extraordinary capability, machine learning models at the same time suffer from more sensitivity as non-semantic features are disturbed. This is particularly the case in tasks heavily interacted with human. 
\item \emph{Utilizable as features}: Classical machine learning is typically designed to employ only the human-perceptible information. Realizing model's capability of processing human-imperceptible information, we envision the opportunities to proactively instruct model to exploit non-semantic feature especially to address the problem of data shortage.
\end{itemize}

\section{Exploiting Adversarial Attack for Benign Applications}\label{sec:3}
Using adversarial attack as instantiation to cultivate non-semantic features, in this section, we discuss how to utilize the above three characteristics to correspondingly design benign applications. The following elaborates each application direction with motivation, justification analysis and prototypes.

\subsection{Adversarial Turing Test}
\subsubsection{Motivation}
\textbf{}

Turing test was initially proposed to examine whether machine’s behavior is indistinguishable from human~\cite{machinery1950computing}, and later developed into reverse Turing test for practical purpose to distinguish between machine and human. The most popular and practical reverse Turing tests are character-based CAPTCHAs ~\cite{naor1996verification} and dialogue-based Turing test~\cite{lowe2017towards}.  However, the recent development of deep learning has significantly advanced the capability of machine in solving both types of reverse Turing tests, e.g., Google duplex demonstrates to crack the dialogue Turing test by creating a chat robot and successfully confounding humans~\cite{leviathan2018google}. It is easy to imagine that with the further development of machine learning, the traditional reverse Turing tests based on intelligence competition will fail to discriminate machine from human. 

Instead of increasing the task complexity to compete between model and human, a straightforward solution for reverse Turing test is to employ their intrinsic different characteristics. As discussed above, non-semantic feature serves as one such characteristic which is exclusive to model and imperceptible to human. With adversarial example disturbing the non-semantic features, we are motivated to design adversarial Turing test by examining the different sensitivities to adversarial perturbation and distinguishing between human and model.

\begin{figure}[t]
\centering

\subfigure[adversarial perturbation distortion.]{
\begin{minipage}[t]{0.5\linewidth}
\centering
\includegraphics[width=\linewidth]{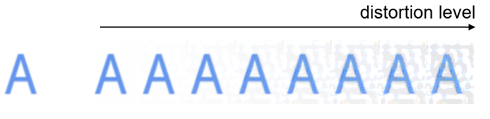}
\end{minipage}%
}%
\subfigure[Gaussion white noise distortion. ]{
\begin{minipage}[t]{0.5\linewidth}
\centering
\includegraphics[width=\linewidth]{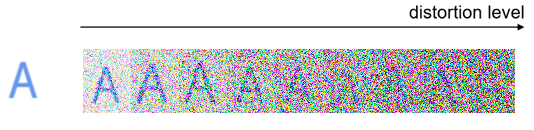}
\end{minipage}%
}%
\quad                 
\subfigure[sensitivity to adversarial perturbation]{
\begin{minipage}[t]{0.5\linewidth}
\centering
\includegraphics[width=\linewidth]{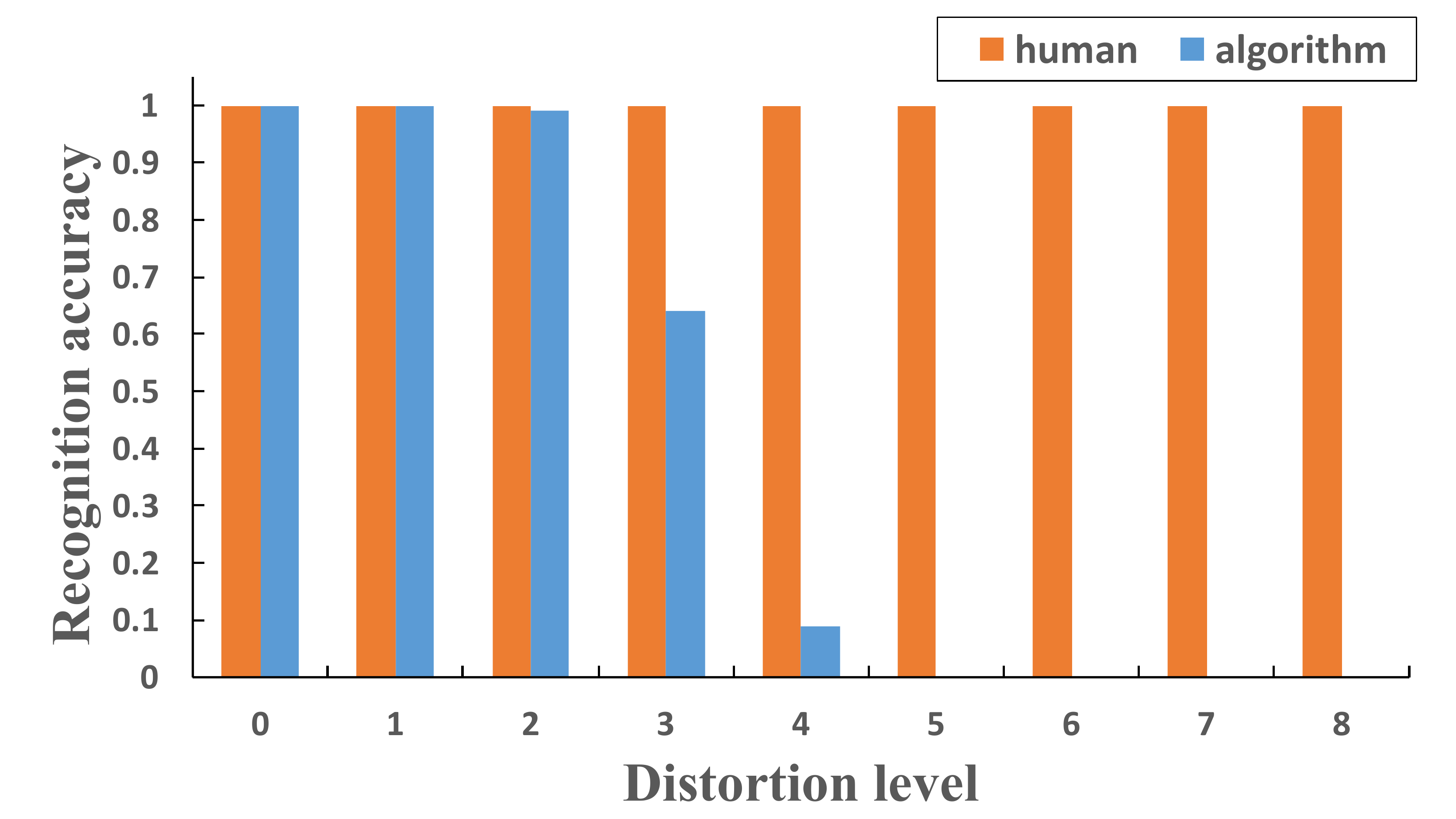}
\end{minipage}
}%
\subfigure[sensitivity to Gaussion noise]{
\begin{minipage}[t]{0.5\linewidth}
\centering
\includegraphics[width=\linewidth]{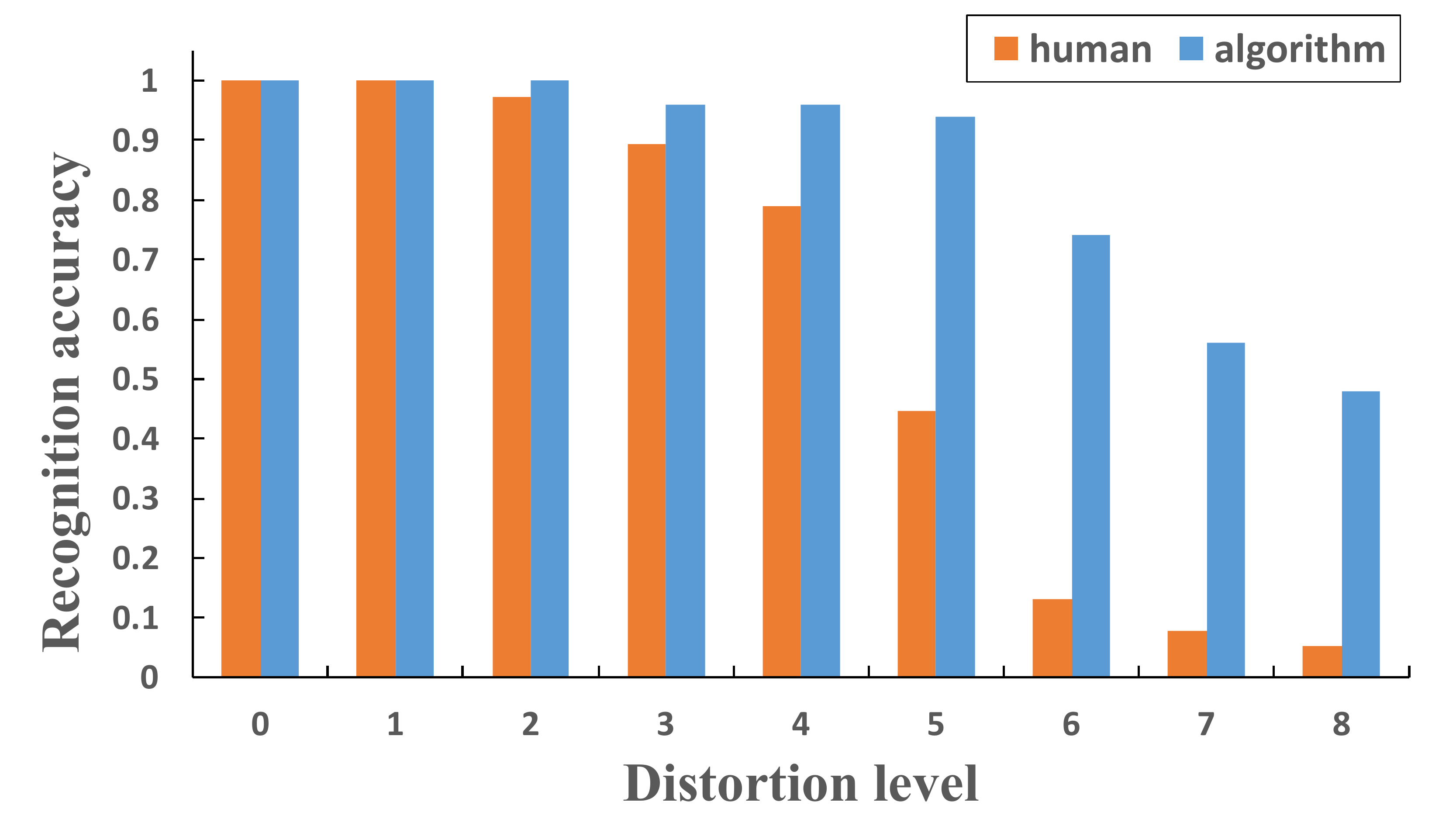}
\end{minipage}
}%
\centering
\label{fig:exp1-3}
  \caption{Different sensitivities of human and model to adversarial and Gaussian distortions.}
\end{figure}

\subsubsection{Justification Analysis}
\paragraph{\textbf{Datasets and settings.}} \vspace{-1mm} We conducted experiments on the modified Alphabet dataset~\cite{krizhevsky2009learning}. The resultant dataset is modified by generating 26 English letters added by random noise, with 1,000 training images and 200 test images for each letter class. We employed a state-of-the-art OCR (Optical Character Recognition) model~\cite{breuel2013high} as experimental model architecture. Adam optimizer is utilized with learning rate of $1e^{-5}$ to train models.\\
\paragraph{\textbf{Observation on the different sensitivities to visual distortions between human and model.}}   \vspace{-1mm}
To verify the different vulnerabilities to adversarial perturbation between human and model, we respectively examine recognition capability of human and model regarding different distortions. Two types of visual distortions are considered specifically. (1) Adversarial perturbation. We employ FGSM~\cite{goodfellow2014explaining} to generate adversarial examples, where one-time perturbation is comprised with step size of 0.02. (2) Gaussian white noise. The added one-time Gaussian white noise follows normal distribution with mean $\mu=0$, variance $\sigma=0.01$ and constant power spectral density. To investigate the tendency of recognition performance with increasing distortions, we crafted $8$ levels of distortions onto the original character images accumulatively: each level corresponds to 5 one-time adversarial perturbations or Gaussian white noises. Examples of derived distorted images at different levels are illustrated in Fig.3 (a) and (b). To examine the recognition capability, we implemented a widely-used OCR~\cite{breuel2013high} regarding the model side. Regarding the human side, we recruited $77$ master workers from Amazon Mechanical Turk. Each subject was asked to recognize $450$ character images with adversarial and Gaussian distortions at different levels.


Fig.3 (c) and (d) illustrate the average recognition accuracies for adversarially and Gaussian distorted CAPTCHAs, respectively. We found human and model show very different sensitivities to the two types of distortions: (1) For adversarially distorted CAPTCHAs, humans are very robust to the adversarial perturbations, while OCR model is highly vulnerable as the distortion level increases. We explain this result as that adversarial perturbations affect the non-semantic features which mislead the algorithm inference while remain imperceptible to human. (2) For Gaussian distorted CAPTCHAs, we observed quite opposite results. While human's recognition accuracy declines sharply as the distortion level increases, OCR model demonstrates relatively stable performance. This is possibly due to the fact that heavy Gaussian noise corrupts the semantic information and thus significantly affects human recognition. However, OCR model manages to recognize the character by employing the non-semantic information. Therefore, this analysis validates that non-semantic features reveal the difference between human and model, showing potential to be applied in designing Turing test.

\subsubsection{Prototype Application and Discussion}
\textbf{}

We have implemented a prototype application by employing adversarially perturbed images to improve character-based CAPTCHAs, which is called robust CAPTCHAs~\cite{zhang2020robust}. Robust CAPTCHAs proposes to exploit the intrinsic difference between human and model on the utilization of non-semantic features. This guarantees the identification of model from human and thus protects against potential cracking without increasing human's burden. In addition to the standard adversarial attack, components of multi-target attack, ensemble adversarial training and differentiable approximation are also proposed to address the characteristics obstructing robust CAPTCHA design, e.g., image preprocessing, sequential recognition, black-box crack. 

With the tendency that traditional Turing test being cracked by machine learning models, researchers are continuously exploring alternative solutions. As the exclusive information of the model, non-semantic feature is justified via the above analysis and prototype application for its feasibility in distinguishing between human and model. We hope this study can shed light on the future studies on exploiting adversarial examples to design novel Turing tests. Moreover, besides the standard forms of Turing test, we envision the increasing necessity of generalized Turing test in more scenarios. In particular, the widespread application of machine learning models in data synthesis~\cite{chesney2019deep} and automated data annotation~\cite{andriluka2018fluid} is giving rise to a considerable amount of model-generated data in the wild. We believe the utilization of non-semantic feature inevitably leaves traces in the generated data, which provides possible solution to identify these model-generated data by carefully examining its reaction to adversarial perturbation.


\begin{table*}[t]
\centering
\caption{ASRs with different adversarial attack methods.}\label{tab:1}
\begin{tabular}{c|c|ccccc} 
\toprule[1pt]
Models & Datasets & FGSM & I-FGSM & MI-FGSM & DI$^2$-FGSM & D-MI$^2$-FGSM\\
\midrule
\multirow{3}*{ArcFace} & LFW & 98.5\% & 99.4\% & 99.4\% & 99.4\% &	99.3\% \\
    & AgeDB-30& 95.8\% & 96.0\% & 96.0\% & 96.0\% &	96.0\% \\
    & CFP-FP       & 92.5\% & 93.7\% & 90.8\% & 93.2\% &	93.4\%\\
\midrule
\midrule
\multirow{3}*{MobileFaceNet}
    & LFW  & 74.5\%	& 86.8\% & 88.2\% &	92.4\% & 89.1\%\\
    & AgeDB-30 & 81.7\%	& 86.3\% & 88.1\% &	90.5\% & 88.6\% \\
    & CFP-FP     & 48.6\%	& 57.2\% & 63.8\% &	68.3\% & 65.0\%\\

\bottomrule[1pt]
\end{tabular}
\end{table*}

\subsection{Rejecting Malicious Model Application}
\subsubsection{Motivation}
\textbf{}

Malicious model application refers to that machine learning model is utilized by hackers to disservice the community. Take face recognition model as example, it has been widely adopted in scenarios like criminal monitoring, security unlock, digital ticket and even face payment. However, there also exist hackers that leverage face recognition models to crawl face images for malicious use. For example, Kneron tested that widely-used face payment systems like AliPay and WeChat can be fooled by masks and face images, and DeepFake creating imitated videos causes severe personal sabotage~\cite{chesney2019deep}. As one of the critical biometric authentication information, the leakage of portrait information will lead to serious security problems. Nevertheless, massive face images are shared onto photo sharing platforms like Facebook and Instagram. Therefore, it is vital to protect privacy from malicious face recognition models when sharing face images online. 

As non-semantic feature is critical to affect model inference while being imperceptible to human, adding adversarial perturbation to face images before sharing is capable of misleading face recognition models while remaining the images' perceived quality. As a result, face privacy can be successfully preserved from being maliciously used without affecting the utility of face image.

\subsubsection{Justification Analysis}

\paragraph{\textbf{Malicious face recognition settings.}}  \vspace{-1mm} We employed two state-of-the-art face recognition methods as the potential malicious models to be rejected. One is ArcFace~\cite{deng2019arcface}, which is widely-used in public Face ID systems. The other is  MobileFaceNet~\cite{chen2018mobilefacenets}, which is commonly used on mobile devices. 
\begin{itemize}
\item[$\bullet$]\textbf{ArcFace} is the popularly utilized face recognition algorithm in public Face ID systems, whose backbone network is Resnet-v2-152 ~\cite{he2016identity}. 
\item[$\bullet$]\textbf{MobileFaceNet} is specifically tailored for high-accuracy real-time face verification using minimal computing resources, whose backbone network is MobileNet-V2 ~\cite{sandler2018mobilenetv2}.
\end{itemize}

We trained the two face recognition models on MS-Celeb-1M dataset~\cite{guo2016ms}, and tested on three datasets of LFW \cite{huang2008labeled}, AgeDB-30 \cite{moschoglou2017agedb}, and CFP-FP \cite{sengupta2016frontal}. With goal of preventing the model from matching the same person, we use the positive pairs of images belonging to the same person for testing: one image as the enrolled image to represent identity, and the other image for synthesizing adversarial example. 

\paragraph{\textbf{Adversarial attack settings.}}  \vspace{-1mm}
We respectively employed FGSM~\cite{goodfellow2014explaining}, I-FGSM\cite{kurakin2016adversarial}, MI-FGSM\cite{dong2018boosting}, DI$^2$-FGSM\cite{xie2019improving} and D-MI$^2$-FGSM\cite{xie2019improving} to generate adversarial examples.  Attack Success Rate (ASR) is selected to evaluate the effectiveness of adversarial attack to mislead the face recognition models:
\begin{equation}
    ASR= \frac{ N_{w/o}-N_{w}}{N_{total}}
\end{equation}
where $N_{w/o}$ and $N_{w}$ denote the number of correctly recognized face images without and with perturbation, respectively. $N_{total}$ denotes the total number of face images. The higher ASR, the better the effectiveness of adversarial attack in rejecting face recognition models.

\paragraph{\textbf{Observation I: rejecting malicious face recognition models.}}   \vspace{-1mm}

We employed the above introduced adversarial attack methods on the two face recognition models of ArcFace and MobileFaceNet. Table~\ref{tab:1} shows the ASRs on the three testing datasets. We can find that: (1) Both for ArcFace and MobileFaceNet, all adversarial attack methods achieve considerable ASRs. The reason for the relative lower ASR of MobileFaceNet is due to its inferior recognition accuracy on original clean images. This verifies the effectiveness of employing adversarial examples to obstruct malicious face recognition models from privacy leaking.  (2) Within each row, ASR consistently increases from left to right corresponding to stronger adversarial attack methods. This encourages the employment of stronger adversarial attacks towards more promising malicious model rejection.

\begin{table}\label{tab-2.C}
\centering
\caption{ASRs on ArcFace using different datasets.}\label{tab:tableexp2-3}
\begin{tabular}{c|ccc} 
\toprule[1pt]
 Datasets & $makeup$  & $adversarial$ & $makeup\_adversarial$\\
\midrule

LFW & 5.1\%	& 98.5\%  & \textbf{98.7\%}\\
 AgeDB-30 & 9.5\% & 95.8\%  & \textbf{99.4\%}\\
CFP-FP & 4.8\% &   92.5\%&\textbf{96.9\%}\\
\bottomrule[1pt]
\end{tabular}
\end{table}

\begin{figure}[t]
  \centering
  \includegraphics[width=\linewidth]{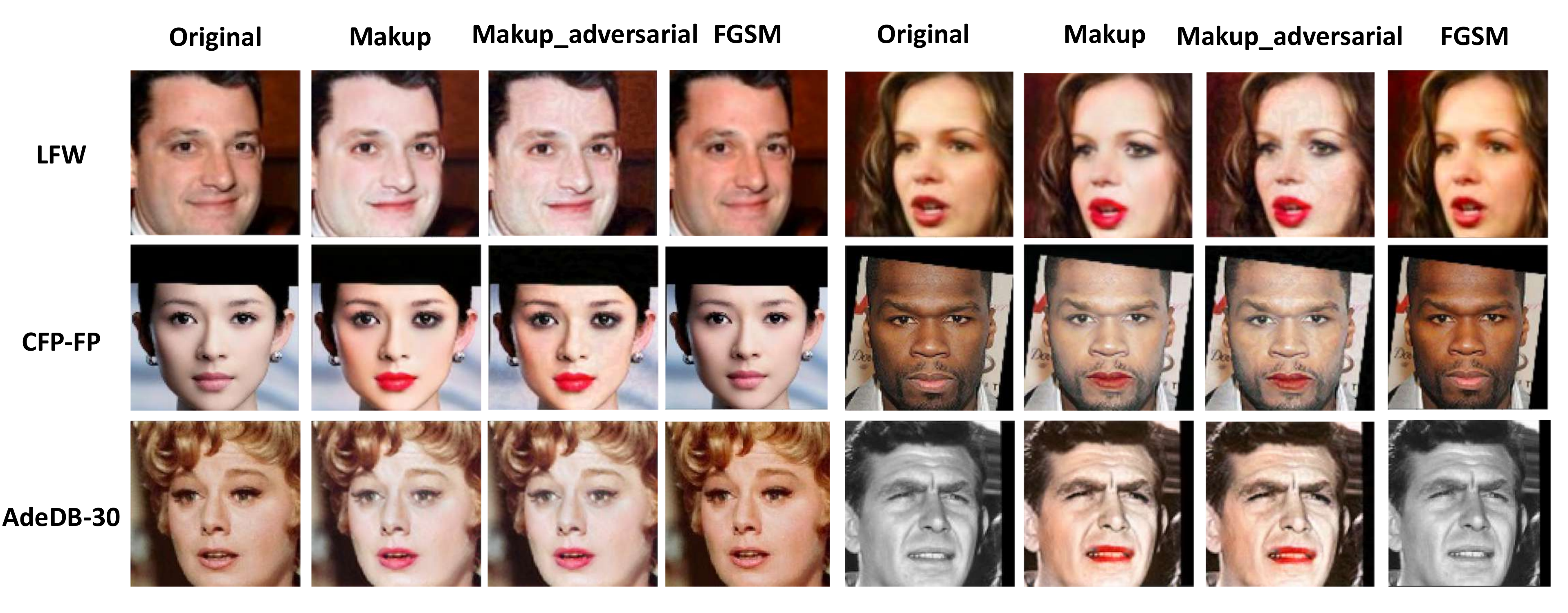}
  \caption{Examples of \emph{makeup}, \emph{adversarial} and \emph{makeup$\_$adversarial} images. }
  \label{fig:exp2-3}
\end{figure}

\paragraph{\textbf{Observation II: compatibility with facial makeup.}}   \vspace{-1mm}
People are increasingly adding makeup to beautify their portrait images before sharing online. In this case, it is necessary for adversarial attacks to remain effective on the images with facial makeup. To verify this, we combined the state-of-the-art facial makeup method PSGAN~\cite{jiang2020psgan} and adversarial attack method FGSM to generate adversarial examples. ASR results on the three datasets are summarized in Table~\ref{tab:tableexp2-3}: \emph{makeup} indicates the images only processed by PSGAN, \emph{adversarial} indicates the images only perturbed by FGSM, and \emph{makeup\_adversarial} indicates the images first processed by PSGAN and then perturbed by FGSM. Example images are illustrated in Fig.~\ref{fig:exp2-3}. We can see that adversarial attack does not undermine the visual perception of makeup image, which guarantees makeup utility as well as privacy-preserving effectiveness. 

From the results we have the following observations: (1) \emph{Makeup} contributes slightly to confusing the face identify (with low yet positive ASR). We suspect that facial makeup corrupts task-relevant semantic information so as to affect the face recognition performance. ~(2) The fact that \emph{Makeup\_adversarial} achieves higher ASR than both \emph{makeup} and \emph{adversarial} demonstrates the compatibility between makeup and adversarial attack. It is interesting in future studies to explore how to effectively integrate the semantic (e.g., makeup) and non-semantic (e.g., adversarial attack) features. 

\subsubsection{Prototype Application and Discussion}
\textbf{}

A prototype application employing adversarial attack to rejecting malicious face recognition models is implemented~\cite{zhang2020adversarial}. This application introduces an end-cloud collaborated adversarial attack framework, which addresses the advanced requirement to guarantee the original image only accessible to users’ own device. This succeeds in avoiding the leakage of face images during transmission to the server. 

It is expected that adversarial attack can be applied in more scenarios of rejecting malicious models. Other privacy-preserving scenarios are like protecting personal identity in surveillance video and phone call. Example applications of information hiding are like adding adversarial perturbation to sensitive information in documentation to prevent from automated information hacking. 


\begin{figure}[t]
  \centering
  \vspace{-3mm}
  \includegraphics[width=0.95\linewidth]{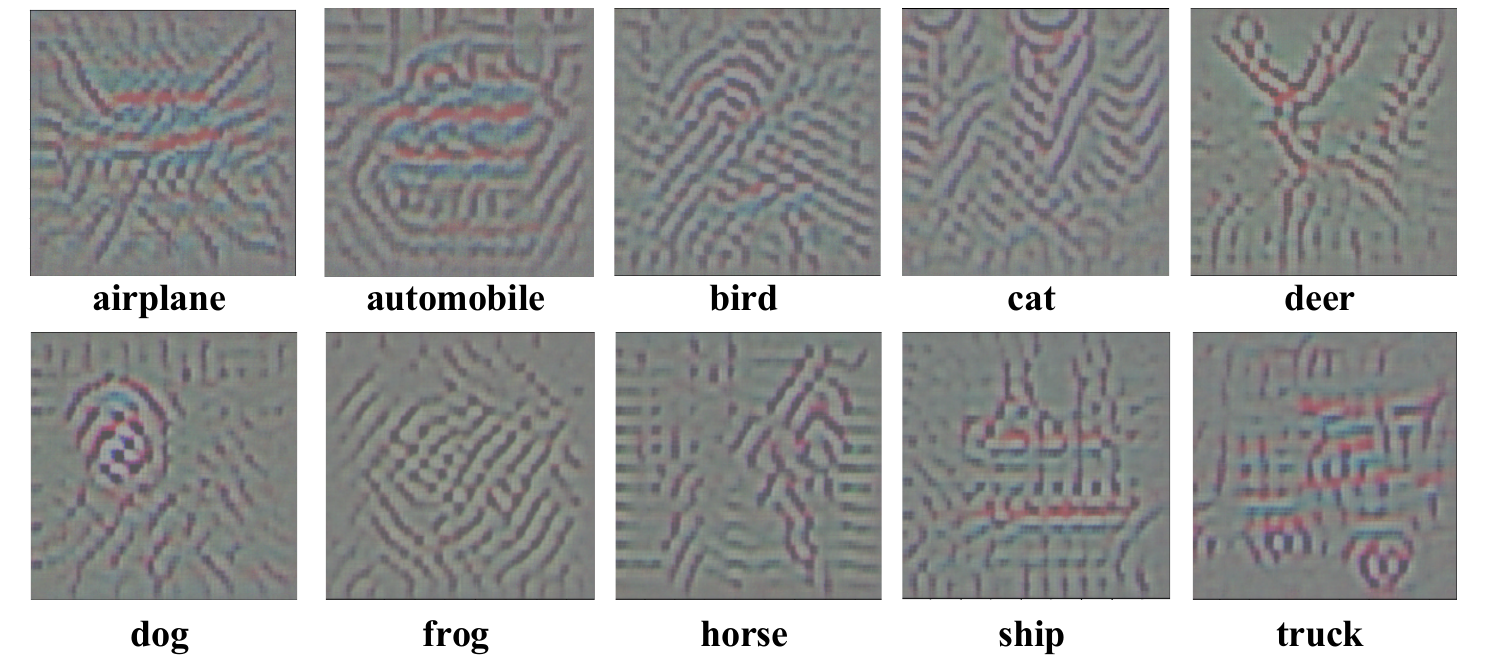}
  \caption{Example UAPs for the 10 categories of CIFAR-10. }
  \label{fig:exp3-1}
  
\end{figure}

\subsection{Adversarial Data Augmentation}
\subsubsection{Motivation}
\textbf{}

The successful application of deep learning is largely due to the adequate annotated data.  However, in many real world applications, it is usually impossible to collect and annotate sufficient training data, e.g., fraud user judgment in finance, abnormal behavior detection in risk control, tumor diagnosis in medicine, etc. This results in the so-called data shortage problem, which gives rise to notorious issues like overfitting~\cite{hawkins2004problem} and model bias~\cite{kirchner2016machine, hardt2016equality}. Typical solutions to address this are few-shot learning from model side and data augmentation from data side. While few-shot learning resorts to extracting as many generalizable features as possible from the limited training data, data augmentation is committed to generating new training data to enhance the original training data. It is worth noting that the above solutions both focus on employing the human-perceptible information, e.g., transferring semantic features from base classes to novel classes in few-shot learning~\cite{sun2019mtl}, augmenting data along semantic directions like image rotation and translation to encourage model to learn more task-relevant semantic features~\cite{wang2019implicit}. 

Inspired from the discussion on model's employment of non-semantic features, we envision the potential of exploiting the non-semantic information to alleviate the data shortage problem. We review that the essence of data shortage lies in the lackage of human-perceptible semantic information and human annotation.  Augmenting data via adversarial attack provides alternative solutions by discovering and employing the non-semantic features for free.

\subsubsection{Justification Analysis}
\paragraph{\textbf{Dataset and settings}}\vspace{-2mm}
We justify the motivation to employ adversarial attack to extract non-semantic features on the CIFAR-10~\cite{krizhevsky2009learning} dataset.  A special type of adversarial perturbation called Universal Adversarial Perturbation (UAP) is employed as tool to analyze the representative non-semantic information used by machine learning models. It is known that standard adversarial attacks are sample-specific, which means different samples even in the same category needs different adversarial perturbations. UAP, however, is sample-independent which misleads most of the input samples for given model and category~\cite{moosavi2017universal}. In this regard, UAP can serve as probe to explore what features the model relies on~\cite{poursaeed2018generative,zhang2020adversarial}. 

\begin{figure}[t]
  \centering
  \includegraphics[width=0.92\linewidth]{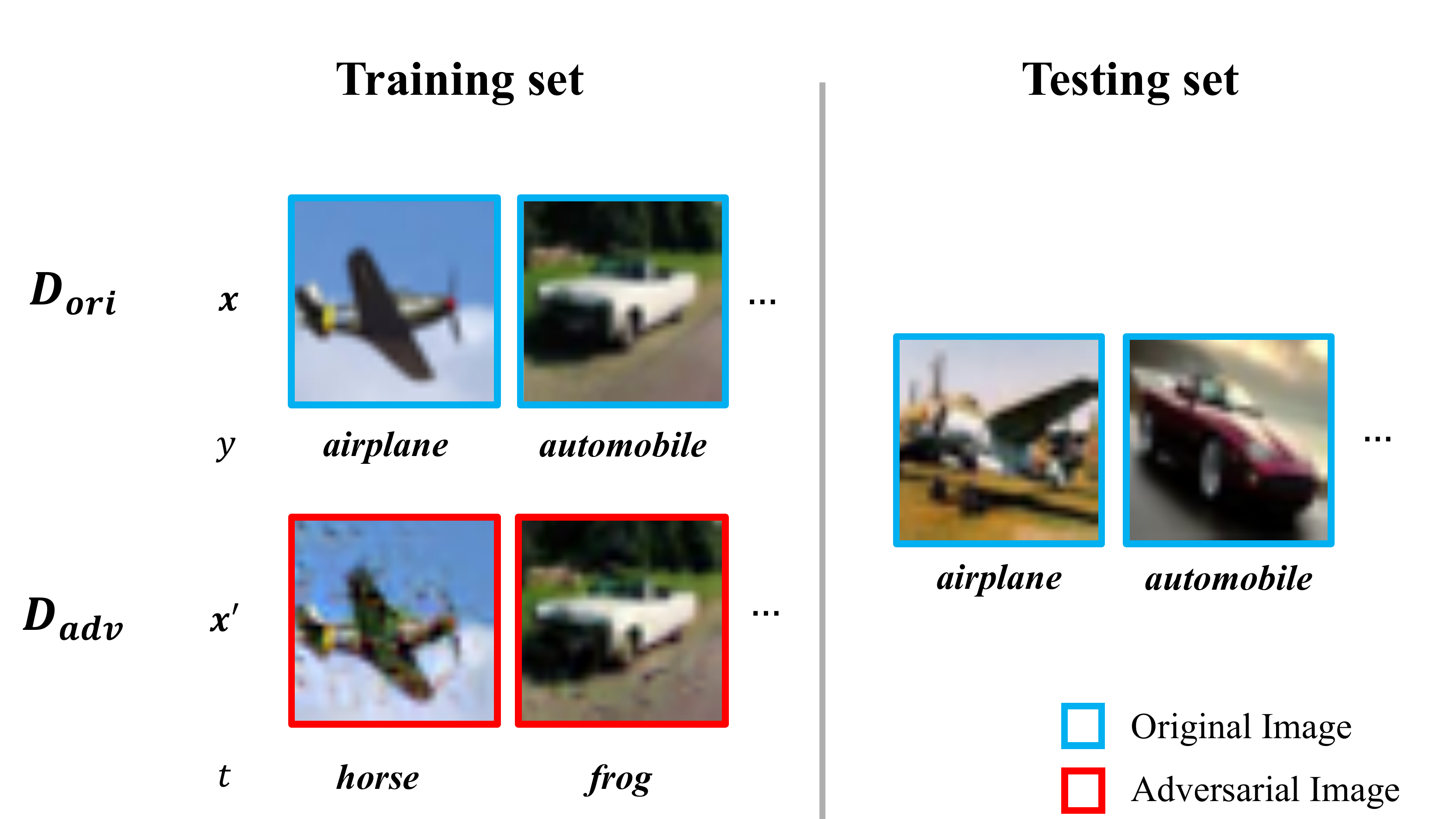}
  \caption{Experimental setting of training set on $D_{ori}$, $D_{adv}$ and testing set.}
  \label{fig:exp3-3-2}
\end{figure}

 

\paragraph{\textbf{Observation I: adversarial attack provides strong feature.}}\vspace{-2mm}
We employed UAP to analyze whether adversarial attack can provide discriminative non-semantic features. Fig.~\ref{fig:exp3-1} shows the generated UAPs for the 10 categories in CIFAR10. While we can hardly perceive semantics related to the target category from the UAPs, the original images are misclassified as the target category once added with the corresponding UAP patterns. This implies that the adversarial perturbation contains very strong features which even surpasses the semantic information in the original image. To exclude the entanglement with original image, we also examined how UAP alone will influence the model inference, i.e., adding UAP to blank image and directly feeding it for inference. We surprisingly observed that model still outputs the attacked target label with 100.00\% confidence. This demonstrates that adversarial attack indeed provides discriminative features both combined with the original information and employed independently.


\paragraph{\textbf{Observation II: adversarial attack provides generalizable feature.}}\vspace{-2mm}
We further analyze whether the extracted features from adversarial attack are generalizable to the unseen data. To address this, we replicated the experiments of \cite{ilyas2019adversarial} and trained models only on adversarially attacked images and then tested on the original images. Specifically, we constructed the adversarially attacked dataset $D_{adv}$ by attacking each image $(\mathbf{x}, y)$ from the original training dataset $D_{ori}$: firstly randomly assigning a target category $t\neq y$, and then adding adversarial perturbation onto original image $\mathbf{x}$ so as to mislead the inference result as $t$. $D_{adv}$ consisting of adversarial images $\mathbf{x'}$ and attacked target label $t$ is used to train classification models for class $t$. We illustrate this experimental setting in Fig.~\ref{fig:exp3-3-2}. It is noted that the images in $D_{adv}$ are linked to the target labels by the non-semantic information instead of the human-perceptible semantic information. 

We trained two classifiers of $f_{ori}$ and $f_{adv}$ respectively on $D_{ori}$ and $D_{adv}$, and examined their performance on the original testing dataset. The derived recognition accuracies for $f_{ori}$ and $f_{adv}$ are $94.4\%$ and $66.3\%$. While $f_{adv}$ does not achieve as good performance as $f_{ori}$, the accuracy of $66.3\%$ still demonstrates the non-trivial contribution of adversarial non-semantic feature. Note that recent study reported only a random guessing result on CIFAR10 by randomizing label $y$ but remaining the original image $\mathbf{x}$~\cite{zhang2016understanding}. The only difference between~\cite{zhang2016understanding} and the above experiment lies in the added adversarial perturbation to $\mathbf{x}$. The fact that the classifier learned from adversarially polluted image $(\mathbf{x'},t)$ succeeds to recognize testing image $(\mathbf{x},y)$ demonstrates that, the extracted non-semantic features from adversarial attack have potential to be generalizable to the original samples. 

\begin{figure}[t]
  \centering
  \includegraphics[width=0.85\linewidth]{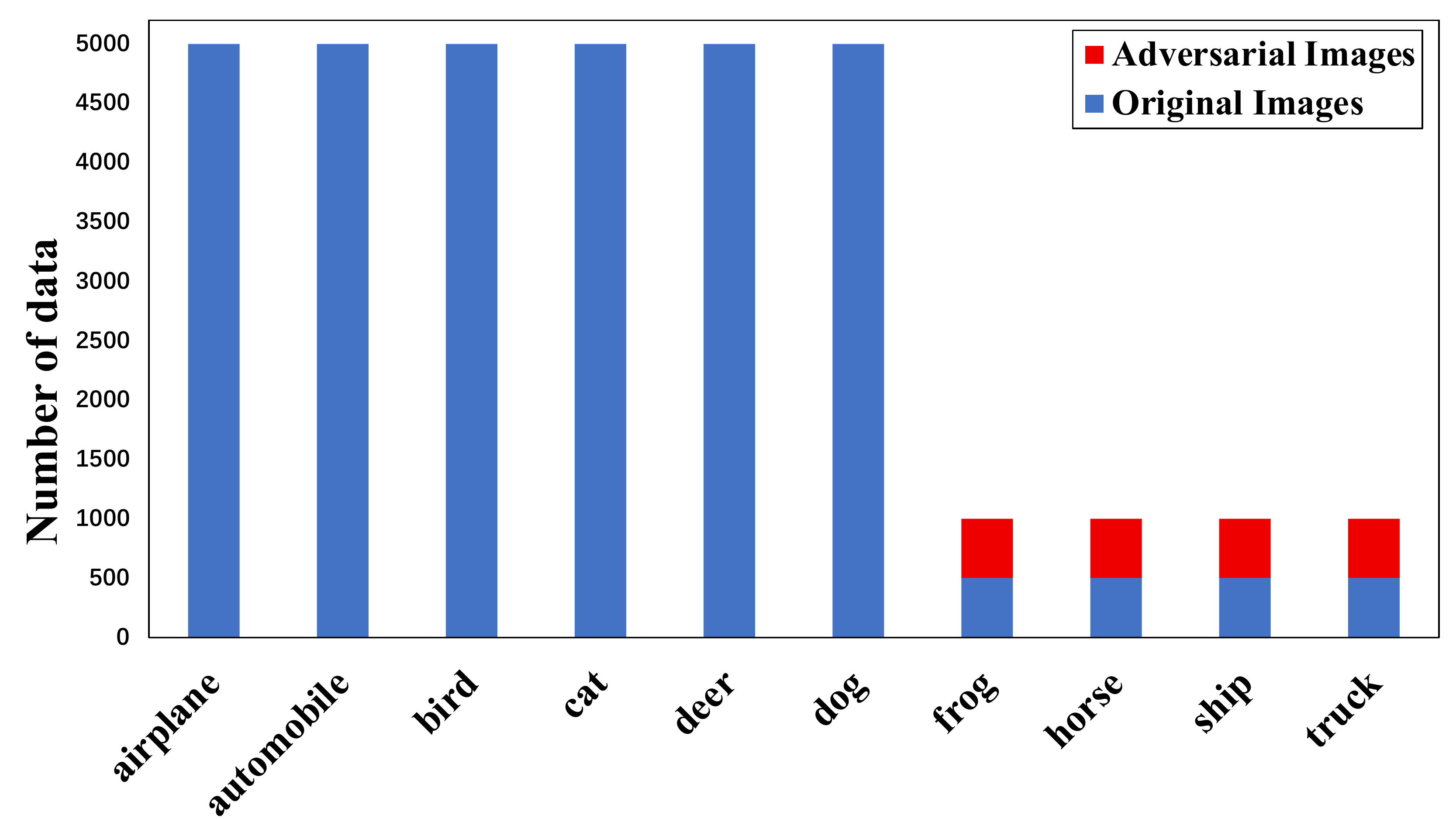}
  \caption{Data distribution of the adversarially augmented training dataset $D_{aug}$.}
  \label{fig:exp3-3a}
\end{figure}

\paragraph{\textbf{Observation III: adversarial attack provides complementary feature.}}  \vspace{-2mm}
In this part, we consider a more practical scenario of exploiting adversarial attack for data augmentation. In particular, a training setting with both original and adversarial images is designed: (1) We first constructed an imbalanced training dataset $D_{imb}$ to imitate data shortage. 4 reduced categories of \emph{frog, horse, ship, truck} are selected which only reserve $10\%$ of the original training data. (2) Next, we expanded $D_{imb}$ to $D_{aug}$ by generating adversarial images by attacking the other 6 categories. Totally another 10\% of adversarial images were obtained for each of the 4 reduced target categories. The data distribution of the adversarially augmented training dataset $D_{aug}$ is illustrated in Fig.~\ref{fig:exp3-3a}.

Similar to the above experiment, we respectively trained two classifiers $f_{imb}$ and $f_{aug}$ on $D_{imb}$ and $D_{aug}$, and tested them on the original testing set. The results for the reduced categories are shown in Table~\ref{tab:8}. We have the following findings: (1) Before augmentation, the lackage of training data significantly affects the performance of classifier $f_{imb}$ to obtain an average accuracy of $49.5\%$. (2) After adversarial data augmentation, the classification accuracy for each category has improved to $54.7\%$. This indicates that adversarial attack provides complementary features to effectively expand the feature pool. 

\begin{table}[t]
\centering
\caption{Classification accuracy on the 4 reduced classes before and after adversarial data augmentation.}\label{tab:8}
\begin{tabular}{c|ccccc} 
\toprule[1pt]
\multirow{2}*{Models} & \multicolumn{5}{c}{Classification accuracy}\\

\cline{2-6}
&frog & horse & ship &truck & \emph{average} \\
\midrule
$f_{imb}$ &44.7\% &42.5\% &56.6\% &54.1\% & \emph{49.5\%} \\
$f_{aug}$ &53.7\% &45.4\% &58.0\% &61.7\% & \textbf{\emph{54.7\%}} \\
\bottomrule[1pt]
\end{tabular}
\end{table}

\subsubsection{Prototype Application and Discussion}
\textbf{}

By utilizing adversarial examples for data augmentation, we implemented a prototype application to solve the model bias problem~\cite{zhang2020towards}. In order to obtain a fair dataset in which the distribution of bias variables is balanced, we apply adversarial attacks to generate examples with minority bias variable as the augmented data. Instead of down-sampling to discard samples or adding additional regularization to sacrifice accuracy as in traditional debiasing solutions, adversarial examples serve as pseudo samples to augment the training dataset and thus simultaneously improve the accuracy and fairness.

Through the above experiments and analysis, we demonstrated that expanding training dataset by adversarial examples is an effective means of data augmentation. While traditional data augmentation can be viewed as exploiting human-consistent prior knowledge to make up for the data shortage, adversarial attack provides a new sight of prior knowledge which is imperceptible to human. We argue that the capability of employing non-semantic information contributes much to the rapid progress of today's machine learning algorithms. However, few work has proactively examined how to exploit it in benign applications. It remains unexplored in many perspectives before  integrating the non-semantic and semantic features for practical usage, e.g., in what cases the models are likely to extract non-semantic features~\cite{zhiyu2022}, how to extract and better employ the non-semantic features, what are the pros and cons of employing non-semantic features, etc.

\section{Conclusion}\label{sec:4}
This work discusses the possibility of exploiting adversarial attack for goodness. Starting with the discussions on human-model disparity, we attribute adversarial example to model's utilization of non-semantic features. In this perspective, adversarial example is not malignant in default but reflecting some interesting characteristics of non-semantic features. We have demonstrated the feasibility of designing benign adversarial attack applications in adversarial Turing test, rejecting malicious model application and adversarial data augmentation. While some applications are ready for practical use, more applications deserve extensive future studies. We hope this study introduces some fresh perspective to consider adversarial attack.  Since there are both negative and positive aspects of adversarial attack, more exploration is required to understand the non-semantic information and control between the unintended risk and new opportunity.

\bibliographystyle{ACM-Reference-Format}
\bibliography{acmart}

\end{document}